\documentclass[final,onefignum,onetabnum]{siuro210301}

\usepackage{lipsum}
\usepackage{amsfonts}
\usepackage{graphicx}
\usepackage{epstopdf}
\usepackage{algorithmic}
\ifpdf
  \DeclareGraphicsExtensions{.eps,.pdf,.png,.jpg}
\else
  \DeclareGraphicsExtensions{.eps}
\fi

\usepackage{enumitem}
\setlist[enumerate]{leftmargin=.5in}
\setlist[itemize]{leftmargin=.5in}

\newsiamremark{remark}{Remark}
\newsiamremark{hypothesis}{Hypothesis}
\crefname{hypothesis}{Hypothesis}{Hypotheses}
\newsiamthm{claim}{Claim}

\headers{Training Implicit Networks for Image Deblurring using JFB}{L. Liu, S. Tong, and L. Zhao}

\title{Training Implicit Networks for Image Deblurring using Jacobian-Free Backpropagation}

\author{Linghai Liu \thanks{Department of Applied Mathematics, Brown University (\email{linghai\_liu@brown.edu}) }
\and Shuaicheng Tong \thanks{Department of Mathematics, University of California, Los Angeles (\email{allentong24@ucla.edu}) }
\and Lisa Zhao \thanks{Department of Statistics, University of California, Berkeley (\email{lisazhao@berkeley.edu}) }
}

\dedication{\small\textit{Project advisor: Samy Wu Fung \thanks{Department of Applied Mathematics and Statistics, Colorado School of Mines}}}

\usepackage{amsopn}

\DeclareMathOperator*{\argmin}{arg\,min}
\graphicspath{{images/}{./}}
\ifpdf
\hypersetup{
  pdftitle={Training Implicit Networks for Image Deblurring using Jacobian-Free Backpropagation},
  pdfauthor={L. Liu, S. Tong, and L. Zhao}
}
\fi

\begin{document}

\maketitle

% REQUIRED
\begin{abstract}
Recent efforts in applying implicit networks to solve inverse problems in imaging have achieved competitive or even superior results when compared to feedforward networks. 
These implicit networks only require constant memory during backpropagation, regardless of the number of layers. 
However, they are not necessarily easy to train. Gradient calculations are computationally expensive because they require backpropagating through a fixed point. 
In particular, this process requires solving a large linear system whose size is determined by the number of features in the fixed point iteration. 
This paper explores a recently proposed method, Jacobian-free Backpropagation (JFB), a backpropagation scheme that circumvents such calculation, in the context of image deblurring problems. 
Our results show that JFB is comparable against fine-tuned optimization schemes, state-of-the-art (SOTA) feedforward networks, and existing implicit networks at a reduced computational cost.
\end{abstract}

\section{Introduction}
Inverse problems consist of recovering a signal, such as an image or a parameter of a partial differential equation (PDE), from noisy measurements, where direct observation of the signal is not possible. As an effort to solve these problems, deep learning techniques have been utilized to acquire high-quality medical images like magnetic resonance imaging (MRI) and computed tomography (CT) \cite{zeng2021review, ahishakiye2021survey, yaqub2022deep}. 

Conventional deep learning approaches for solving inverse problems use deep unrolling \cite{aggarwal2018modl,chen2018learn,monga2019algo, wu2019computationally, liu2019deep, liang2020deep, chun2020photo}, which utilizes a fixed number of iterations usually chosen heuristically. A deep network is ``unfolded" into a wider and shallower network, where each layer is split into multiple sub-layers. While this method allows the network to learn complex patterns in the input data, it suffers from overfitting and the well-known vanishing gradient problem \cite{hochreiter1998vanishing}, not to mention the lack of flexibility compared to other network structures \cite{gilton2021deep, ongie2020deep}. Moreover, they are challenging to train due to memory constraints in backpropagation. Another line of work, feed-forward denoising convolutional neural networks \cite{zhang2017beyond,zhang2017learning,zhang2018ffdnet,tian2019enhanced}, use deep convolutional neural networks (CNNs) to learn the residuals between the ground truth images and noisy observations instead of directly reconstructing the clean underlying images. These end-to-end models are not trained for a particular forward model, so they may require large amounts of labeled data for training \cite{gilton2021deep, ongie2020deep}. 

Recently, deep equilibrium models (DEQs) were proposed \cite{bai2019deep, bai2020multiscale, winston2020monotone, gurumurthy2021joint, kawaguchi2021on, ramzi2023test, ramzi2023shine}. DEQs use implicit networks with weight-tying, input-injected layers that propel the dynamics of latent space representation by sharing the input across layers. Training involves backpropagating through a fixed point of the layer using implicit differentiation, where the number of layers can be deemed infinite. This feature allows implicit networks to save memory significantly since there is no need to save any intermediate values on the backpropagation graph. Despite yielding fixed memory costs and matching performances of other state-of-the-art (SOTA) models, DEQs are still very expensive to train because backpropagation requires solving a Jacobian-based linear system at every gradient evaluation \cite{ongie2020deep}. 
To this end, a Jacobian-Free Backpropagation (JFB) approach was recently introduced to avoid solving the linear system during training \cite{fung2022jfb}.

The theory of JFB allows us to replace the Jacobian matrix with the identity under certain conditions. JFB not only maintained a fixed memory cost but also avoided the substantial computational cost of Jacobian-based methods while ensuring a descent direction \cite{fung2022jfb}. It has performed well in image classification tasks \cite{fung2022jfb}, computational tomography \cite{heaton2021learn,heaton2021feasibility, heaton2023explainable}, traffic routing \cite{heaton2021learn}, and finding the shortest paths \cite{mckenzie2023faster}. A variation of JFB, where the inverse Jacobian was approximated as a perturbed identity matrix and falls back to JFB when the norm between the approximation and the true inversion grows beyond a threshold \cite{ramzi2023shine}, also proved to be successful in image classification. In this paper, we investigate JFB's effectiveness in training implicit networks for inverse problems arising from image deblurring \footnote{Access the GitHub repository at \url{https://github.com/lliu58b/Jacobian-free-Backprop-Implicit-Networks}}.  

\section{Mathematical Background}
\label{sec:main}

\subsection{Problem Setup}
We have $N$ noisy blurred images $\{ d_i \}_{i=1}^{N} \subseteq \mathbb{R}^n$ that we refer to as \textit{measurements}. The underlying \textit{original images}, denoted as $\{ x_i \}_{i=1}^N \subseteq \mathbb{R}^n$, are hidden from the experimenter(s). We use the model: 
\begin{equation}
\label{forward-model}
    d = \mathcal{A} x + \boldsymbol{\varepsilon}
\end{equation}
where the forward operator $\mathcal{A}$ is a mapping from signal space of original images to measurement space and $\boldsymbol{\varepsilon} \in \mathbb{R}^m$ is a noise term that models measurement errors. In this work, we deal with image deblurring, where the forward process is a Gaussian blur. The value of each pixel in a measurement is a weighted average of its neighborhood under a Gaussian kernel with discretized weights of a 2-dimensional Gaussian density, plus noise generated from the hypothesized measurement process. 

\subsection{Traditional Optimization for Deblurring}
A natural idea is to apply $\mathcal{A}^{-1}$ to~\eqref{forward-model} and obtain $x^\ast = \mathcal{A}^{-1} (d - \varepsilon)$ when $\mathcal{A}$ is invertible, which is the case in denoising ($\mathcal{A} = I$) and deblurring ($\cal A$ is the Toeplitz matrix of a convolution operator). This can amplify the noise and result in really bad reconstructions when $\mathcal{A}$ is ill-conditioned. Therefore, we estimate the true image $x^*$ by formulating a regularized optimization problem that minimizes the difference between the reconstructed image and the observed image: 
\begin{equation}
\label{optim-setup}
    x^\ast = \argmin_{x \in \mathbb{R}^n} \ \frac{1}{2}|| \mathcal{A}x - d ||^2_{L^2} + \lambda R(x)
\end{equation}
where $\lambda > 0$ is a tunable parameter, $R(x)$ is a regularizer chosen based on common practice \cite{golub1999tikhonov} or potentially learned from given data \cite{alberti2021learning, haber2003learning, heaton2022wasserstein, afkham2021learning}. We can solve~\eqref{optim-setup} by applying the gradient descent algorithm, with a common choice of $R(x) = \frac{1}{2} ||x||^2_{L^2}$. However, we observe in our numerical experiments that results are not ideal (low SSIM values for reconstructed images). To this end, we explore the use of implicit networks in this work.

\subsection{Implicit Networks and Challenges}
% What are implicit networks and how are they trained?
Implicit networks \cite{el2021implicit} are newly proposed models that also leverage the dataset $\{(d_i, x_i)\}^N_{i=1}$ and are capable of representing a wide range of feedforward models. The idea is to find a fixed point for their weight-tying layers, modeled as a non-linear function $T(\cdot)$, and map it to the inference space. We formulate our implicit network as follows: 
\begin{equation}
\label{implicit-network}
  \begin{array}{l}
    \text{[Equilibrium equation]} \quad T_\Theta (x) = x\\
    \text{[Prediction equation]} \quad x^\ast (d) = x
  \end{array}
\end{equation}
The iteration to find the fixed point reads:
\begin{equation}
\label{iteration}
    x^{k+1}_i = x^{k}_i - \eta \left(\nabla_x || \mathcal{A}x^k_i - d_i||_{L^2}^2 + S_\Theta (x^k_i) \right) := T_\Theta (x_i^k), \quad k = 0, 1, \ldots, K-1
\end{equation}
where $\eta >0$ is step size, $x_i^0$ is the initial ``guess'', $K$ is the number of iterations (layers) in our neural network $\mathcal{N}_\Theta(\cdot)$, and $S_\Theta (\cdot)$ is a trainable network containing all the weights of $\mathcal{N}_\Theta(\cdot)$. Note that the [Equilibrium equation] is satisfied as $K \to \infty$ when $T_\Theta (\cdot)$ is a contraction. The output of the network is $\mathcal{N}_\Theta (d):= x^\ast(d)$, given by the [Prediction equation]. This iterative scheme is called \texttt{DE-GRAD} \cite{gilton2021deep}.

To update $\Theta$, we perform implicit differentiation on the equilibrium equation in (\ref{implicit-network})
$$ \frac{d x^\ast}{d \Theta} = \frac{d T_\Theta (x^\ast)}{d x^\ast} \frac{d x^\ast}{d \Theta} + \frac{\partial T_\Theta (x^\ast)}{\partial \Theta} \overset{\text{rearrange terms}}{\Longrightarrow} \left( I - \frac{d T_\Theta (x^\ast)}{d x^\ast}\right) \frac{d x^\ast}{d \Theta} =  \frac{\partial T_\Theta (x^\ast)}{\partial \Theta}$$
and substitute $\frac{d x^\ast}{d \Theta}$ into the gradient descent scheme after setting up a loss function $\ell$:
\begin{equation}
\label{update-rule}
     \Theta \leftarrow \Theta - \alpha \frac{d \ell}{d x^\ast} \left( I - \frac{d T_\Theta (x^\ast)}{d x^\ast}\right) ^{-1} \frac{\partial T_\Theta (x^\ast)}{\partial \Theta}
\end{equation}
where $\alpha > 0$ is the learning rate and $(I - \frac{d T_\Theta (x^\ast)}{d x^\ast})$ is the Jacobian matrix $\mathcal{J}$. This update rule is costly due to the need to invert $\mathcal{J}$, which motivates the search for other ways to speed up the backpropagation process.

\section{Related Works}
\label{sec:related}

\subsection{Deep Unrolling for Inverse Problems}
Deep Unrolling \cite{aggarwal2018modl,chen2018learn,monga2019algo, wu2019computationally, liu2019deep, liang2020deep, chun2020photo} is used to unpack deep neural networks, whose black-box nature hinders their training and interpretation. Intuitively, each iteration is ``unfolded" into smaller layers and then concatenated to form a deep network. Therefore, these neural networks can be interpreted as an optimization problem as in~\eqref{optim-setup}, with a fixed number of $K$ iterations upon initialization, and the regularizer $R(x)$ can be parametrized to adaptably regularize the training process to minimize the loss of each estimate $x^k$. Deep Unrolling has achieved successful results in other inverse problems in imaging, such as low-dose CT \cite{wu2019computationally}, light-field photography \cite{chun2020photo}, blind image deblurring \cite{li2019deep}, and emission tomography \cite{mehranian2020emission}. In our setup, this method corresponds to~\eqref{iteration}, with a finite $K$ being fixed. From prior numerical experiments \cite{gilton2021deep, ramzi2023test}, $K$ is eventually a relatively small, handcrafted number as a result of fine-tuning and limitations in time and space for both training and inference. 

\subsection{Implicit Networks}
Deep Equilibrium Models (DEQs), a type of implicit networks, were proposed \cite{bai2019deep, bai2020multiscale, winston2020monotone, gurumurthy2021joint, kawaguchi2021on, gilton2021deep, geng2021training}. The advantage of DEQ lies in that it requires less memory because it uses a weight-tied, input-injected design, where it only has one layer of actual weights and the original input is fed into each of the identical layers. It solves the fixed-point and uses implicit differentiation to calculate the gradient for backpropagation. On the other hand, SOTA deep feed-forward networks such as Deep Unrolling have memory issues since they store intermediate values while iterating through each network layer.

\section{Proposed Methodology}
\label{sec:methodology}
The convergence criterion of $T_\Theta (\cdot)$ in~\eqref{iteration} (\texttt{DE-GRAD}) is discussed in more detail in \cite{gilton2021deep}, where $S_\Theta - I$ needs to be $\epsilon-$Lipschitz to ensure that $T_\Theta (\cdot)$ is contraction with parameter $\gamma \in [0, 1)$. 
We propose using Jacobian-free Backpropagation (JFB) \cite{fung2022jfb}, a  recently-introduced algorithm that updates $\Theta$ at a lower computational cost. The idea is to circumvent the Jacobian calculation in (\ref{update-rule}) by replacing $\mathcal{J}$ with the identity $I$, leading to an approximation of the true gradient: 
\begin{equation}
    p_\Theta = \frac{d \ell}{d x^\ast} \frac{\partial T_\Theta (x^\ast)}{\partial \Theta}
\end{equation}
which is still a descending direction for the loss function $\ell$ with more constraints on $T_\Theta$ \cite{fung2022jfb}. Approximating $\mathcal{J}$ with $I$ is equivalent to taking the first term of the Neumann series $\left(I - \frac{d T_\Theta (x^\ast)}{d x^\ast} \right)^{-1} = \sum_{k=0}^\infty \left(\frac{d T_\Theta (x^\ast)}{d x^\ast}\right)^k$.
Some researchers adopt the first several terms for finer approximations while training Jacobian-based implicit networks \cite{geng2021training}. Note the difference here compared to implicit networks is that we are inverting the identity matrix rather than the Jacobian $\mathcal{J}$ in~\eqref{update-rule}. Like in other works, we used Anderson acceleration \cite{walker2011anderson} to facilitate the process of finding fixed points for the mapping $T_\Theta(\cdot)$ while keeping \texttt{torch.no\_grad()}. After finding the root $x^\ast$, we resume the gradient tape and output $T_\Theta(x^\ast) = x^\ast$ as an input of loss $\ell$. Then, we use PyTorch to calculate $\frac{d\ell}{dx^\ast} \frac{\partial T_\Theta(x^\ast)}{\partial \Theta} = \frac{d\ell}{dx^\ast} \frac{\partial x^\ast}{\partial \Theta}$, which is $\mathcal{O}(n^2)$ because we fixed the dimension of parameters once training starts. Even though JFB is not always performing the steepest descent, the JFB step is much less costly to calculate, which lowers its overall cost while ensuring a descent direction. In contrast, implementing other models with implicit differentiation requires multiplying by $\frac{d \ell}{dx^\ast}$ and $\frac{\partial T_\Theta(x^\ast)}{\partial \Theta}$, which is also $\mathcal{O}(n^2)$, to the left and right of the inverse $\mathcal{J}^{-1}$ \cite{geng2021training}. The complexity of these update rules is augmented mainly by inverting the Jacobian, which is hard to build explicitly and requires solving a large linear system. 

\begin{algorithm}
\label{training}
\caption{Learning with JFB}
\begin{algorithmic}
\REQUIRE Implicit network $N_\Theta (\cdot)$ with weight-tying layers $T_\Theta (\cdot)$. Set learning rate $\alpha > 0$. 
\FOR {measurement-truth pair $(d, x)$ in training set}
\STATE Find fixed point: $x^\ast = T_\Theta (x^\ast; d)$ with \texttt{torch.no\_grad()}
\STATE Output: $\mathcal{N}_\Theta (d) = T_\Theta (x^\ast)$
\STATE Calculate loss: $\ell(x^\ast, x)$
\STATE Update: $\Theta \leftarrow \Theta - \alpha \frac{d \ell}{d x^\ast} \frac{\partial T_\Theta (x^\ast)}{\partial \Theta}$
\ENDFOR
\end{algorithmic}
\end{algorithm}

\section{Experimental results}
\label{sec:experiments}

We mimic the format in \cite{gilton2021deep} while implementing our JFB approach. That is, we perform our experiments on the same dataset and use the same quality measures for image reconstruction. 
\begin{itemize}
    \item \textbf{Data}: We use a subset of size 10,000 of the CelebA dataset \cite{liu2015deep}, which contains around 200,000 centered human faces with annotations. Among the subset of 10,000, 8,000 images are used for training and the rest are left for validating and testing purposes. 
    
    \item \textbf{Preprocessing}: Each image is resized into $128 \times 128$ pixels with 3 channels (RGB) and normalized into range $[0,1]$ with mean $\frac{1}{2}$ for each channel. The blurred images are generated using Gaussian blurring kernels of size $5 \times 5 $ with variance $1$. The measurements are then crafted by adding white Gaussian noise with standard deviation $\sigma = 10^{-2}$ to the blurred images. 

    \item \textbf{Network Architecture}: We use a convolutional neural network (CNN) structure with 17 layers. Except for the first and last CNN layer, each intermediate layer is followed by batch normalization and element-wise ReLU activation function. We also applied spectral normalization to each CNN layer to make sure that the mapping is Lipschitz continuous with Lipschitz constant no greater than 1. 
    
    \item \textbf{Training}: With the aforementioned data and preprocessing specifics, the training strictly follows Algorithm \hyperref[training]{1}, where $T_\Theta (\cdot)$ is the same as defined in~\eqref{iteration}. As part of $T_\Theta$, the trainable network $S_\Theta (\cdot)$ is also pretrained as practiced in \cite{gilton2021deep} to observe an improvement in reconstruction. 
    
    \item \textbf{Visualization}: We first visualize the average training and validation loss per image over the number of epochs in Fig. \hyperref[loss-plot]{1}, running on a sample of 2000 images, with an 80-20 training-validation split. 
    \begin{figure}[ht]
        \label{loss-plot}
        \centering
        \includegraphics[width=0.5\linewidth]{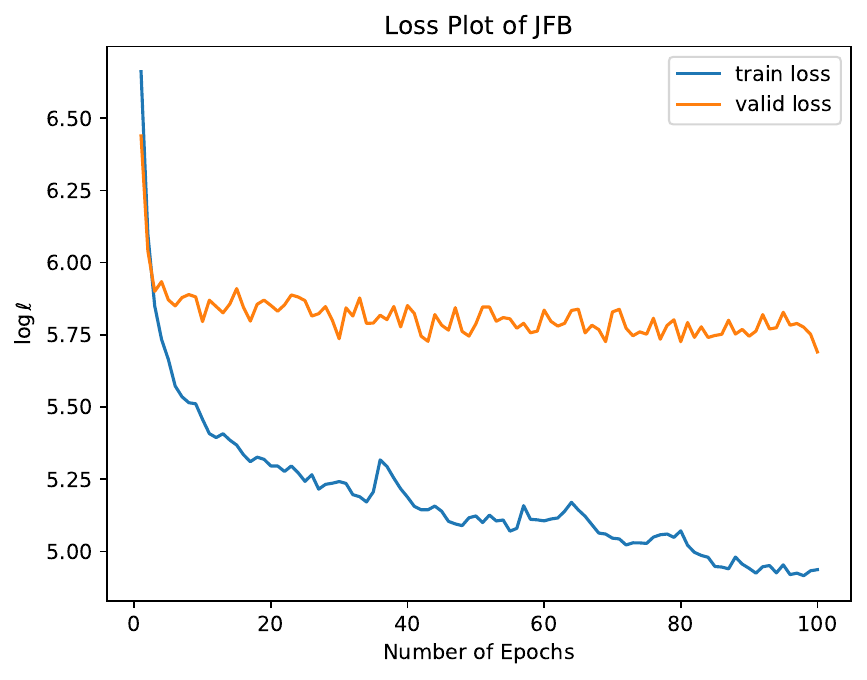}
        \caption{Training and Validation losses of the DE-GRAD model with JFB}
    \end{figure}
    
We see that as the number of epochs increases, the training loss decreases, while the validation loss remains relatively high. For this proof of concept, we are using only a subset of our dataset, so it is hard for the model to generalize well to the validation images. However, the validation loss still shows a decreasing trend. Also, the JFB algorithm only promises a descent direction for the loss function, rather than a technique that learns the structure of the data distribution. Nevertheless, the comparable PSNR and SSIM values reported in Table \ref{table::PSNR-SSIM} were calculated with a JFB model that achieved an MSE of about 100 per image $(\log 100 \approx 4.605)$. 
    \begin{table}[ht]
        \centering
        \begin{tabular}{|c|c|c|c|c|}
        \hline
          \begin{tabular}{l}
            Ground Truth
          \end{tabular} &
          \begin{tabular}{l}
        \includegraphics[width=0.12\linewidth]{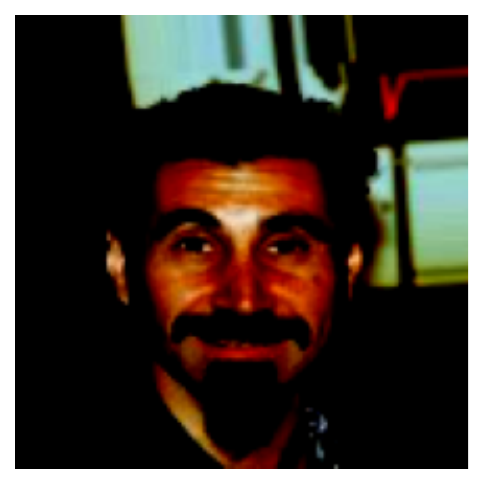}
          \end{tabular} &
          \begin{tabular}{l}
         \includegraphics[width=0.12\linewidth]{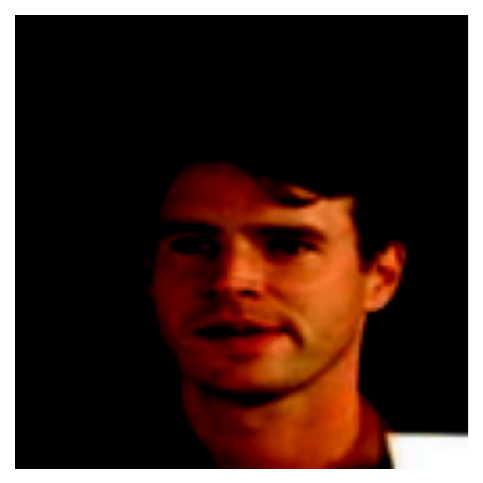}
          \end{tabular} &
          \begin{tabular}{l}
         \includegraphics[width=0.12\linewidth]{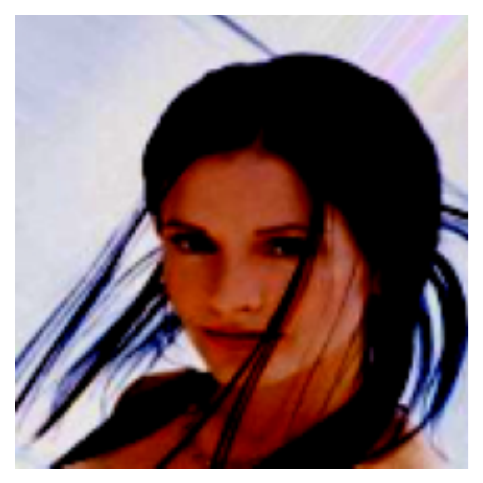}
          \end{tabular}&
          \begin{tabular}{l}
         \includegraphics[width=0.12\linewidth]{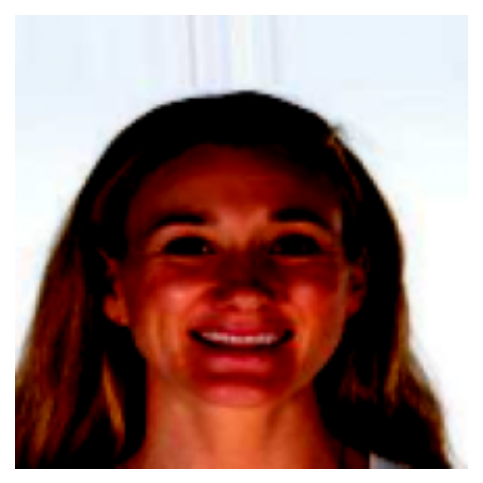}
          \end{tabular}
         \\
          \hline
          \begin{tabular}{l}
            Noisy Blurred Image
          \end{tabular} &
          \begin{tabular}{l}
        \includegraphics[width=0.12\linewidth]{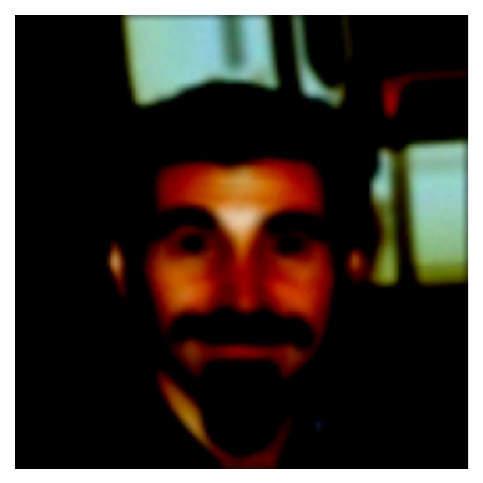}
          \end{tabular} &
          \begin{tabular}{l}
         \includegraphics[width=0.12\linewidth]{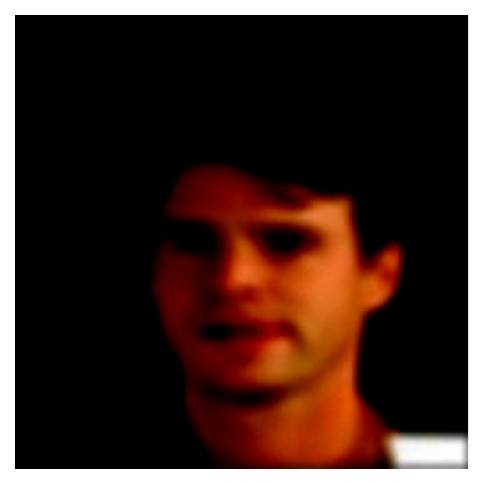}
          \end{tabular} &
          \begin{tabular}{l}
         \includegraphics[width=0.12\linewidth]{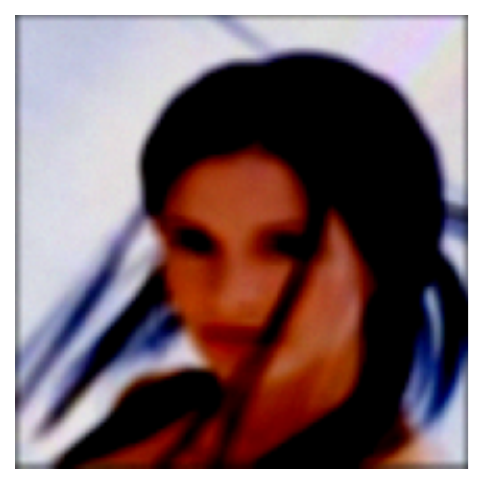}
          \end{tabular}&
          \begin{tabular}{l}
         \includegraphics[width=0.12\linewidth]{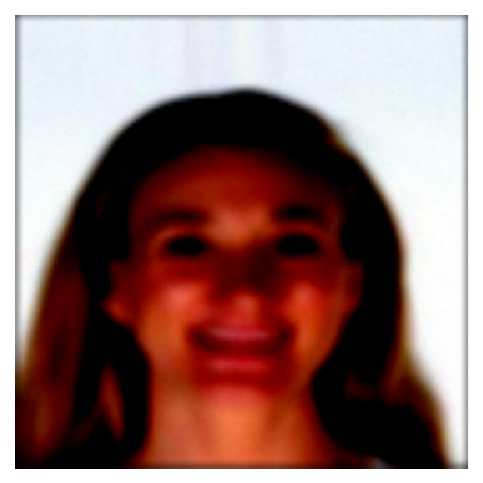}
          \end{tabular}
          \\
          \hline
          \begin{tabular}{l}
            Direct Inverse
          \end{tabular} &
          \begin{tabular}{l}
        \includegraphics[width=0.12\linewidth]{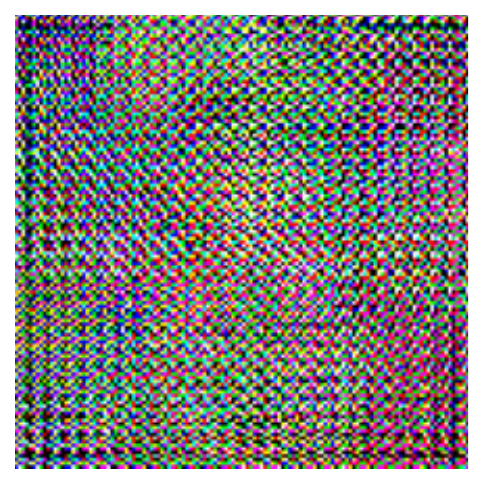}
          \end{tabular} &
          \begin{tabular}{l}
         \includegraphics[width=0.12\linewidth]{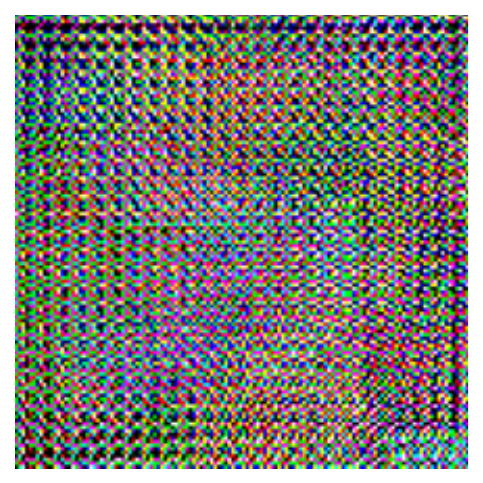}
          \end{tabular} &
          \begin{tabular}{l}
         \includegraphics[width=0.12\linewidth]{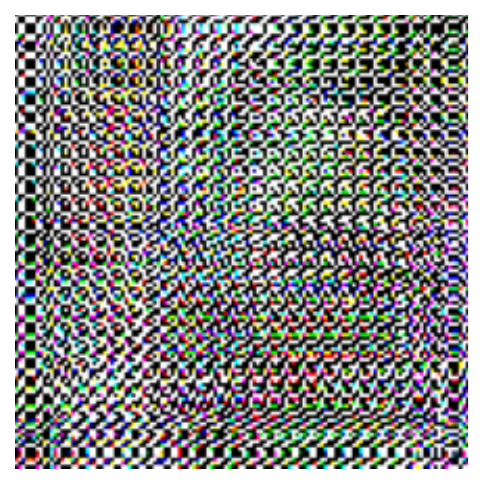}
          \end{tabular}&
          \begin{tabular}{l}
         \includegraphics[width=0.12\linewidth]{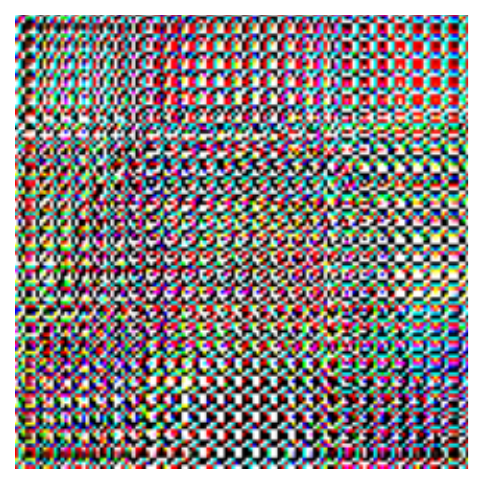}
          \end{tabular}
          \\
          \hline
          \begin{tabular}{l}
            Gradient Descent
          \end{tabular} &
          \begin{tabular}{l}
        \includegraphics[width=0.12\linewidth]{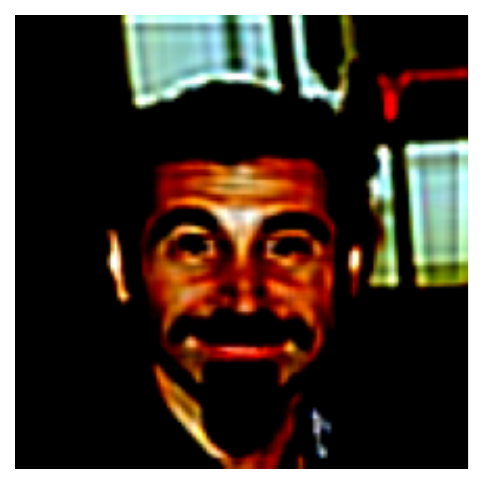}
          \end{tabular} &
          \begin{tabular}{l}
         \includegraphics[width=0.12\linewidth]{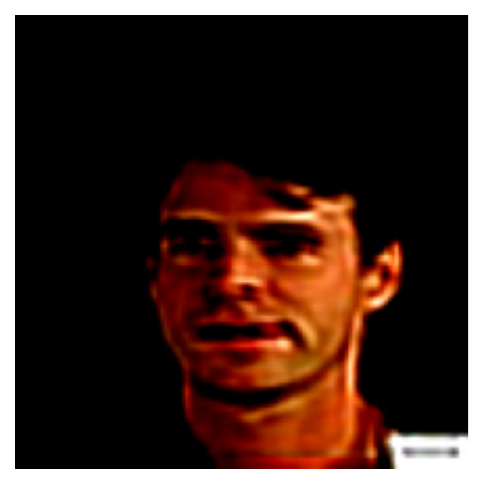}
          \end{tabular} &
          \begin{tabular}{l}
         \includegraphics[width=0.12\linewidth]{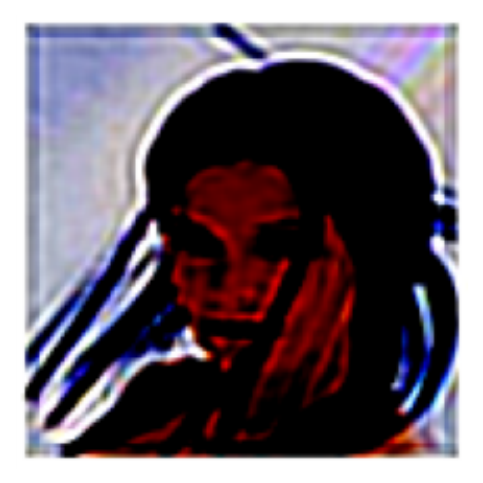}
          \end{tabular}&
          \begin{tabular}{l}
         \includegraphics[width=0.12\linewidth]{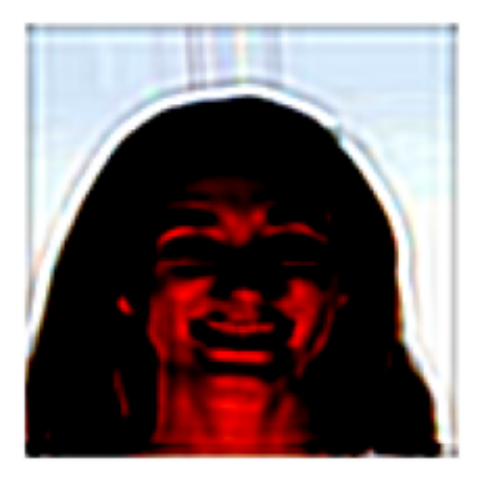}
          \end{tabular}
          \\
          \hline
          \begin{tabular}{l}
            JFB
          \end{tabular} &
          \begin{tabular}{l}
        \includegraphics[width=0.12\linewidth]{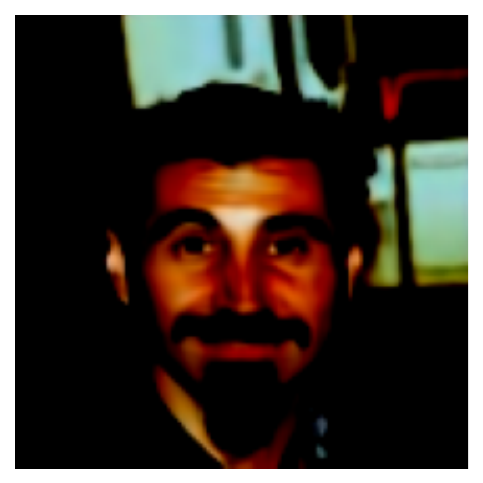}
          \end{tabular} &
          \begin{tabular}{l}
         \includegraphics[width=0.12\linewidth]{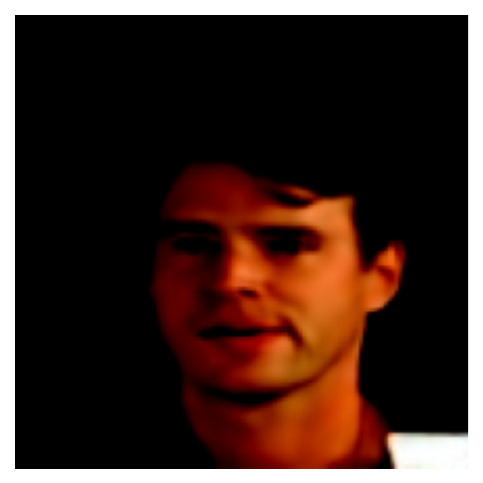}
          \end{tabular} &
          \begin{tabular}{l}
         \includegraphics[width=0.12\linewidth]{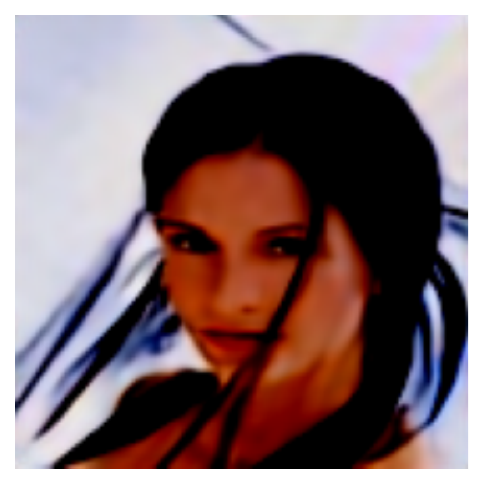}
          \end{tabular}&
          \begin{tabular}{l}
         \includegraphics[width=0.12\linewidth]{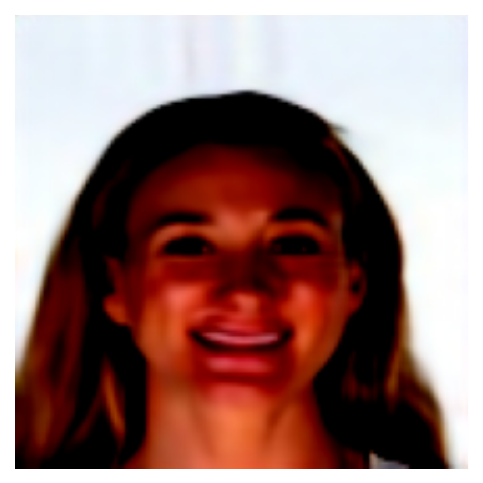}
          \end{tabular}
          \\
          \hline
        \end{tabular}
\caption{Visualization of JFB}
\label{tb:visual}
\end{table}

    We then visualize examples from the \texttt{DE-GRAD} model trained using JFB in Table \ref{tb:visual}, using the same four images for each row. Ground Truths are the images in CelebA dataset resized to $128 \times 128$ pixels. Noisy Blurred Images are generated by a $5 \times 5$ Gaussian kernel mentioned above. Direct Inverse images are the results of applying the inverse of $\mathcal{A}$ on blurred images, which is defined in~\eqref{forward-model}. Gradient Descent images are obtained by applying gradient descent using mean squared error (MSE) as the loss function. We enforce early stopping so that the results are not corrupted. JFB images are obtained using Noisy Blurred images as inputs with JFB-trained weights for the \texttt{DE-GRAD} model. 
    
    \item \textbf{Comparison of Quality}: We compare our results obtained from JFB with other state-of-the-art methods using total variation (TV), standard deep neural networks, and DEQ. The metrics we use to assess the quality of reconstructed images are peak-signal-to-noise ratio (PSNR, a positive number, best at $+\infty$) and structural similarity index measure (SSIM, also positive, best at $1$) \cite{hore2010image}. The data in Table \hyperref[compare]{2} are calculated on the testing dataset of size 2000. Although we have not achieved results better than DEQs, which finds the true gradient in a complicated manner in~\eqref{update-rule}, we currently observe results that are competitive with other techniques.
    \begin{table}[ht]
    \label{table::PSNR-SSIM}
        \label{compare}
            \centering
            \begin{tabular}{|c|c|c|c|c|}
            \hline
            & Total Variation & Plug-n-Play \cite{venkatakrishnan2013plug} & Deep Equilibrium \cite{gilton2021deep} & JFB (Ours) \\
            \hline
            PSNR & 26.79 & 29.77 & 32.43 & 26.88 \\ 
            \hline
            SSIM & 0.86 & 0.88 & 0.94 & 0.91 \\
            \hline
            \end{tabular}
        \caption{Comparison across Models}
    \end{table}
    
     \item \textbf{Comparison of Time and Complexity}: We compare the complexity per gradient/step computation using JFB and Jacobian-based backpropagation on the CelebA dataset. Here, $n = 128 \times 128 \times 3 = 49152$ after preprocessing.
     Importantly, we note that JFB requires a Jacobian matrix-vector product per sample, leading to complexity $\mathcal{O}(n^2)$. Jacobian-based backpropagation, however, also requires the inverse of $\mathcal{J}$, i.e., solving a linear system as shown in~\eqref{update-rule}.
     
     Na\"ively using PyTorch for gradient tracking and taking the inverse of the Jacobian during our experiments depletes all possible RAM (more than 32 Gigabytes). Hence, we use the conjugate gradient method to solve the linear system in~\eqref{update-rule}. The idea is to let $w = \frac{d \ell}{d x^\ast}\mathcal{J}^{-1}$, and solve $w\mathcal{J}\mathcal{J}^T = \frac{d \ell}{d x^\ast}\mathcal{J}^T$ instead of $w\mathcal{J} = \frac{d \ell}{d x^\ast}$. 

     In our experiments, we fix a batch of 16 images in the CelebA dataset, run both the Jacobian-based update and JFB for 20 repetitions, and then record the average time needed in seconds for each parameter update. The dimensions of the images are ``Image Size" by ``Image Size" by 3 channels, where ``Image Size" varies from 16 to 128 at an increment of 16. 
     \begin{figure}[ht]
        \label{time-plot}
        \centering
        \includegraphics[width=0.55\linewidth]{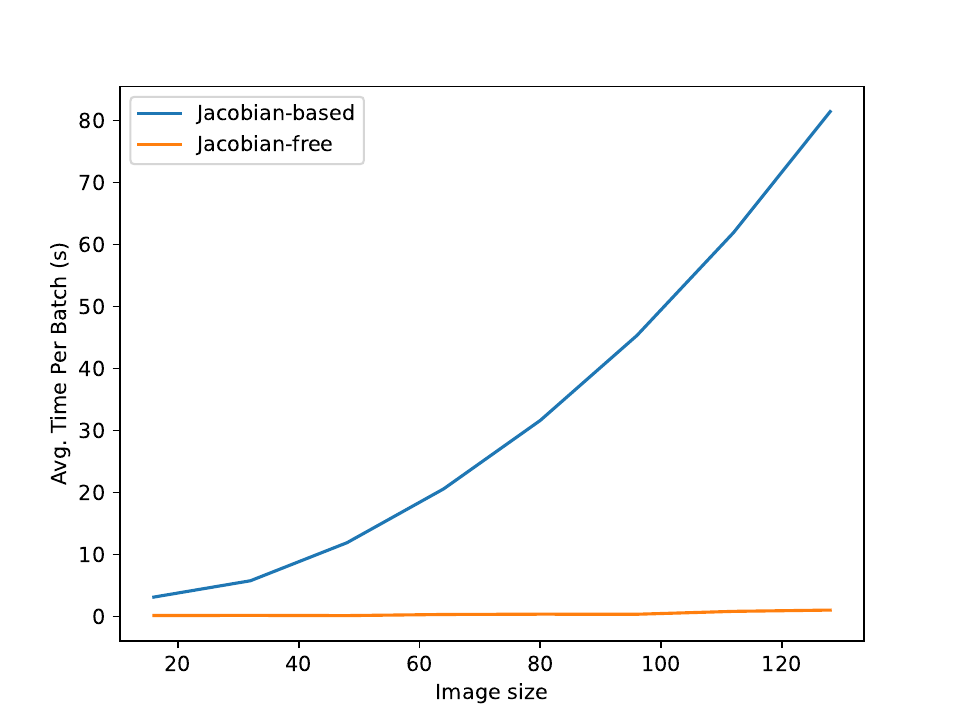}
        \caption{Computation Time per Batch for Jacobian-based method v.s. JFB}
    \end{figure}
    
    Figure~\ref{time-plot} shows that while maintaining comparable image reconstruction quality measured by PSNR and SSIM, JFB algorithm is faster and easier to implement with auto-differentiation libraries such as Tensorflow or PyTorch.
\end{itemize}

\section{Conclusions}
\label{sec:conclusions}

In this paper, we explored Jacobian-free Backpropagation for implicit networks with applications in image deblurring. Our approach recovers images rather effectively across the testing dataset. Moreover, JFB is competitive with other state-of-the-art methods whose hand-crafted parameters have been fine-tuned across all stages. We also demonstrated the advantage of JFB in terms of its computational complexity and easiness of implementation in practice. Future work involves application to other inverse problems like denoising \cite{mueller2012linear, rudin1992nonlinear}, geophysical imaging \cite{haber2014computational, fung2019multiscale, fung2019uncertainty, kan2021pnkh}, and more.

\section{Acknowledgements}
We sincerely thank the guidance of our mentor, Dr. Samy Wu Fung, and other mentors at Emory University for the opportunity.
We also thank Emory University for providing the GPUs that made our numerical experiments possible.
This work was partially funded by National Science Foundation awards DMS-2309810 and DMS-2051019.

\bibliographystyle{siamplain}
% \bibliography{references}

\end{document}